\documentclass[11pt,onecolumn]{IEEEtran}
\ifCLASSINFOpdf
\else
\fi
\usepackage{amsmath,graphicx,amssymb,url,epstopdf,mathtools}
\usepackage{etoolbox}
\usepackage{flushend}
\usepackage[multiple]{footmisc}

\DeclarePairedDelimiter\ceil{\lceil}{\rceil}

\begin{document}
%
\title{On Hyperspectral Classification in the Compressed Domain}
%
%
%

\author{Mohammad~Aghagolzadeh, 
        Hayder~Radha
\thanks{M. Aghagolzadeh and H. Radha are with the Department
of Electrical and Computer Engineering, Michigan State University, East Lansing,
MI, 48823 USA e-mail: \{aghagol1, radha\}@msu.edu.}
\thanks{This work was supported by NSF Awards CCF-1117709 and CCF-1331852}
}

\maketitle

\begin{abstract}
In this paper, we study the problem of hyperspectral pixel classification based on the recently proposed architectures for compressive whisk-broom hyperspectral imagers without the need to reconstruct the complete data cube. A clear advantage of classification in the compressed domain is its suitability for real-time on-site processing of the sensed data. Moreover, it is assumed that the training process also takes place in the compressed domain, thus, isolating the classification unit from the recovery unit at the receiver's side. We show that, perhaps surprisingly, using distinct measurement matrices for different pixels results in more accuracy of the learned classifier and consistent classification performance, supporting the role of information diversity in learning. 
\end{abstract}


%
\IEEEpeerreviewmaketitle

\section{Introduction}
\label{sec:intro}

Recently, there has been a surge toward compressive architectures for hyperspectral imaging and remote sensing \cite{spm14_cs}. This is mainly due to the increasing amount of hyperspectral data that is being collected by high-resolution airborne imagers such as NASA's AVIRIS\footnote{\url{http://aviris.jpl.nasa.gov}} and the fact that a large portion of data is discarded during compression or during feature mining prior to learning \cite{hi_review}. It has been noted in \cite{fow14} that many of the proposed compressive architectures are based on the spatial mixture of pixels across each frame and correspond to physically costly or impractical operations while most existing airborne hyperspectral imagers employ scanning methods to acquire a pixel or a line of pixels at a time. To address this issue, practical designs of compressive \textit{whisk-broom} and \textit{push-broom} cameras were suggested in \cite{fow14}. In this work, we tackle the problem of hyperspectral pixel classification based on compressive whisk-broom sensors; i.e. each pixel is measured at a time using an individual random measurement matrix. Extension of the presented analysis for the compressive push-broom cameras is straightforward. 

To set this work apart from existing efforts that have also focused on the problem of classification from the compressive hyperspectral data, such as \cite{fow09}, we must mention two issues with the typical indirect approach of applying the classification algorithms to the \textit{recovered} data: $a$) the sensed data cannot be decoded at the sender's side (airborne device) due to the heavy computational cost of compressive recovery, making on-site classification infeasible, $b$) the number of measurements (per pixel) may not be sufficient for a reliable signal recovery. It has been established that classification in the compressed domain would succeed with far less number of random measurements than it is required for a full data recovery \cite{svm_cs}. However, the compressive framework of \cite{svm_cs} corresponds to using a fixed projection matrix for all pixels which limits the \textit{measurement diversity} that has been promoted by several recent studies for data recovery and learning \cite{fow09j,globsip13,few}.

\begin{figure}[t]
\begin{minipage}[b]{.32\linewidth}
  \centering
  \centerline{\includegraphics[width=\linewidth]{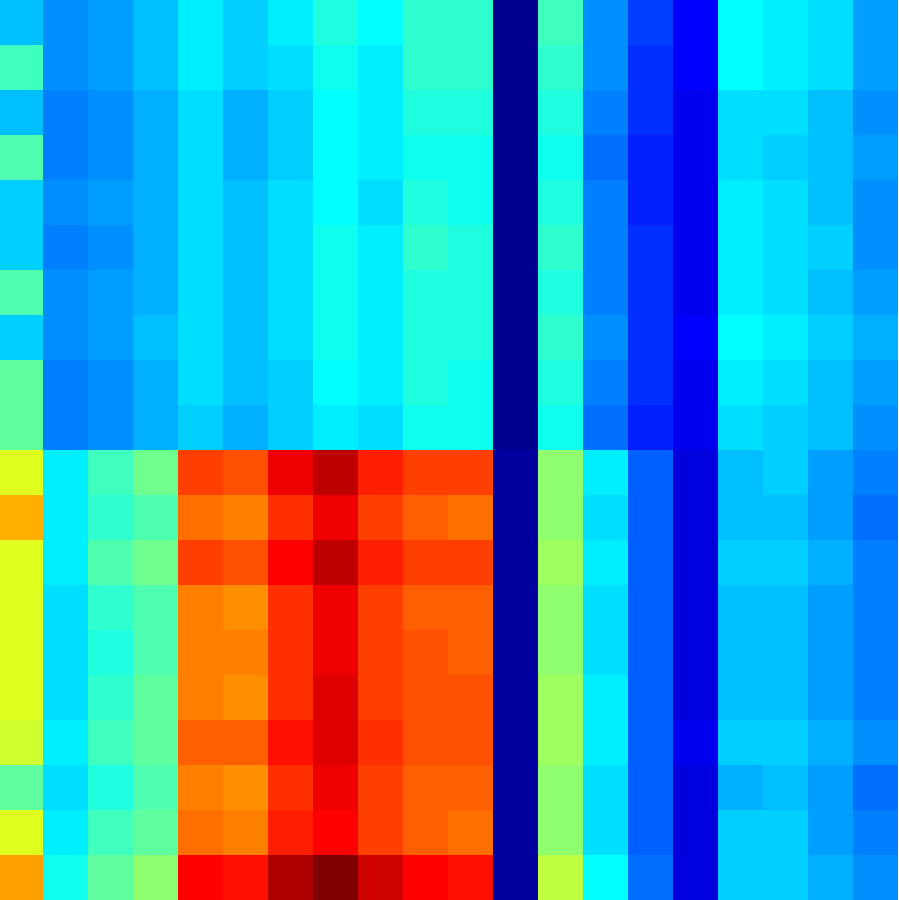}}
  \centerline{Complete data}
\end{minipage}
\hfill
\begin{minipage}[b]{0.32\linewidth}
  \centering
  \centerline{\includegraphics[width=\linewidth]{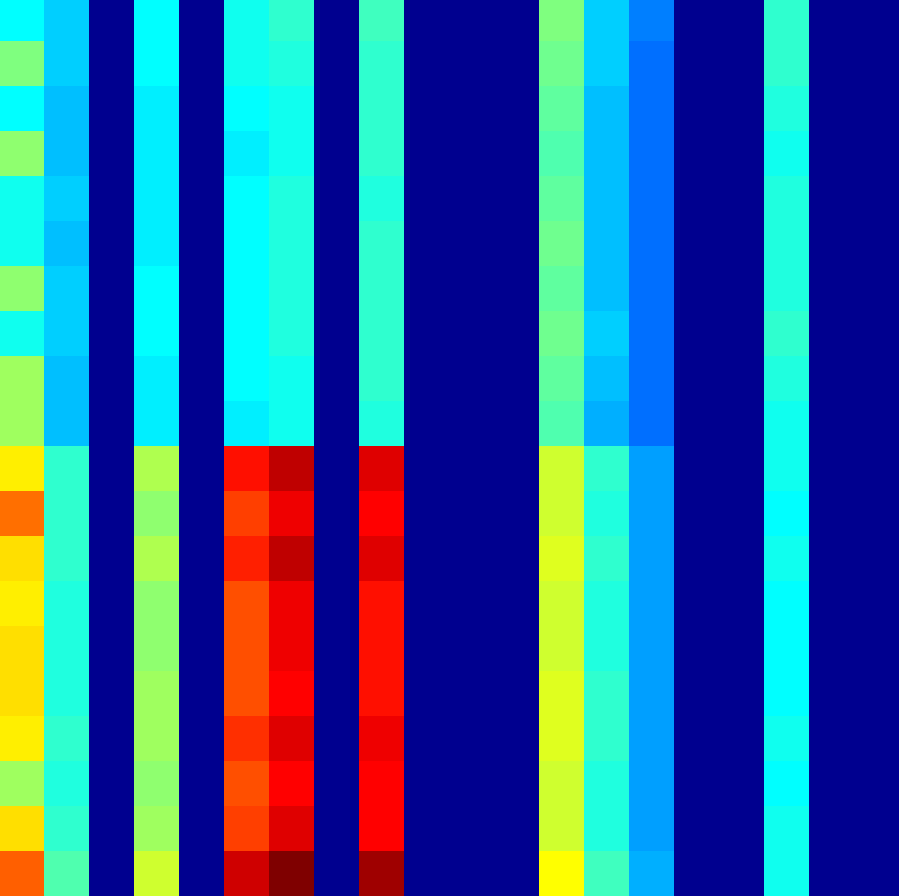}}
  \centerline{FCA-sensed data}
\end{minipage}
\hfill
\begin{minipage}[b]{0.32\linewidth}
  \centering
  \centerline{\includegraphics[width=\linewidth]{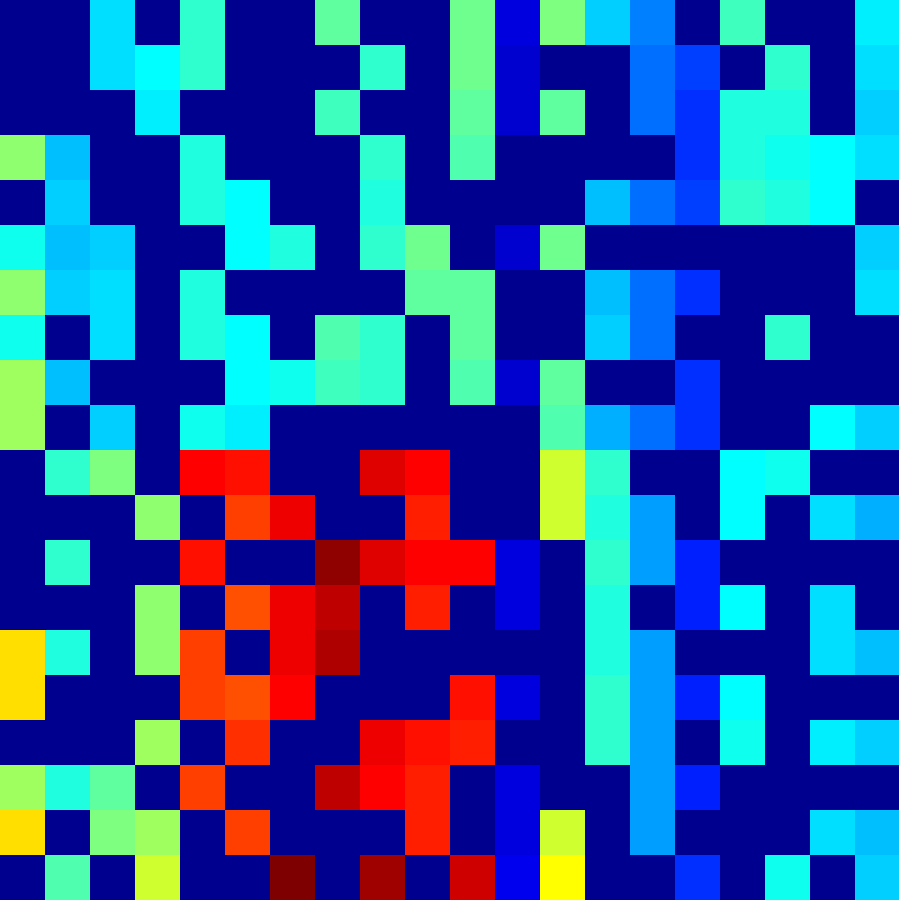}}
  \centerline{DMD-sensed data}
\end{minipage}
\caption{FCA-based versus DMD-based sensing. Here, rows represent pixels and columns represent spectral bands.}
\label{fig:S}
\end{figure}

Rather than devising new classification algorithms, this work is focused on studying the relationship between the camera's sensing mechanism, namely the employed random measurement matrix, and the common Support Vector Machine (SVM) classifier. It must be emphasized that the general problem of classification based on compressive measurements has been addressed for the case where a fixed measurement matrix is used \cite{sp_cs,svm_cs}. However, our aim is to study the impact of \textit{measurement diversity} on the learned classifier. In particular, we investigate two different sensing mechanisms that were introduced in \cite{fow14} \footnote{For more details regarding the physical implementation of compressive whisk-broom sensors, we refer the reader to \cite{fow14} which illustrates conceptual schematics of whisk-broom and push-broom cameras.}:
\begin{itemize}
\item[{\bf 1)}] {\bf FCA-based sensor:} A Fixed Coded Aperture (FCA) is used to modulate the dispersed light before it is collected at the linear sensor array. This case corresponds to using a fixed measurement matrix for each pixel and a low-cost alternative to the DMD system below.
\item[{\bf 2)}] {\bf DMD-based sensor:} A Digital Micromirror Device (DMD) is used to modulate the incoming light according to an arbitrary pattern that is changed for each measurement. Unlike the previous case, DMD adds the option of sensing each pixel using a different measurement matrix. Both cases are illustrated in Figure \ref{fig:S}.
\end{itemize}

SVM has been shown to be a suitable classifier for hyperspectral data \cite{hi_review}. Specifically, we employ an efficient linear SVM classifier with the exponential loss function that gives a smooth approximation to the hinge-loss. To train the classifier in the compressed domain, we must \textit{sketch} the SVM loss function using the acquired measurements for which we employ some of the techniques developed in \cite{sp_cs}. Furthermore, given that the sketched loss function gives a close approximation to the true loss function and that the learning objective function is smooth, it is expected that the learned classifier is close to the ground-truth classifier based on the complete hyperspectral data (which is unknown). As it has been discussed in \cite{r21}, recovery of the classifier is of independent importance in some applications. 

This paper is organized as follows. In the Section \ref{sec:two} we present the learning algorithm that gets the compressive measurements as input and produces a linear pixel classifier in the signal domain. Section \ref{sec:simo} contains the simulation results and their analysis. We conclude the paper in Section \ref{sec:conclusion}.

\section{Problem Formulation and the Proposed Framework}
\label{sec:two}

\subsection{Overview of SVM for spectral pixel classification}
\label{subsec:problem}

In a supervised hyperspectral classification task, a subset of pixels are labeled by a specialist who may have access to the side information about the imaged field such as being physically present at the field for measurement. The task of learning is then to employ the labeled samples for tuning the parameters of the classification machine to predict the pixel labels for a field with similar material compositions. Note that, for subpixel targets, an extra stage of spectral unmixing is required to separate different signal sources involved in generating a pixel's spectrum \cite{r10}. For simplicity, we assume that the pixels are homogeneous (consist of single objects). 

Recall that most classifiers are inherently composed of binary decision rules. Specifically, in multi-categorical classification, multiple binary classifiers are trained according to either One-Against-All (OAA) or One-Against-One (OAO) schemes and voting techniques are employed to combine the results \cite{r11}. In a OAA-SVM classification problem, a decision hyperplane is computed between each class and the rest of the training data, while in a OAO scheme, a hyperplane is learned between each pair of classes. As a consequence, most studies focus on the canonical binary classification. Similarly in here, our analysis is presented for the binary classification problem which can be extended to multi-categorical classification. 

In the linear SVM classification problem, we are given a set of training data points (corresponding to hyperspectral pixels) $x_j\in\mathbb{R}^d$ for $j=1,2,\dots,n$ and the associated labels $z_j\in \{-1,+1\}$. The inferred class label for $x_j$ is $\mbox{sign}(x_j^T w - b)$ that depends on the classifier $w\in \mathbb{R}^d$ and the bias term $b\in \mathbb{R}$. The classifier $w$ is the normal vector to the affine hyperplane that divides the training data in accordance with their labels. When the training classes are inseparable by an affine hyperplane, \textit{maximum-margin soft-margin} SVM is used which relies on a \textit{loss function} to penalize the amount of misfit. For example, a widely used loss function is $\ell(r)=\left(\mbox{max}\{0,1-r\}\right)^p$ with $r=z_j(x_j^T w-b)$. For $p=1$, this loss function is known as the hinge loss, and for $p=2$, it is called the squared hinge loss or simply the quadratic loss. The optimization problem for soft-margin SVM becomes\footnote{\textit{Discussion}: Similar results can be obtained using the dual form. Recent works have shown that advantages of the dual form can be obtained in the primal as well \cite{r12}. As noted in \cite{r12}, the primal form convergences faster to the optimal parameters $(w^*,b^*)$ than the dual form. For the purposes of this work, it is more convenient to work with the primal form of SVM although the analysis can be properly extended to the dual form.}
\begin{equation}
(w^*,b^*)=\arg\min_{w,b} \frac{1}{n}\sum_{j=1}^n\ell(z_j(x_j^T w-b)) + \frac{\lambda}{2} \|w\|^2_2
\label{eq:svm_primal_soft}
\end{equation} 
In this paper, we use the smooth exponential loss function, which can be used to approximate the hinge loss while retaining its margin-maximization properties \cite{exploss}:
\begin{equation}
\ell(z)=e^{-\gamma z}
\label{eq:exp_loss}
\end{equation}
where $\gamma$ controls the smoothness. We use $\gamma=1$.

\subsection{SVM in the compressed domain}

Let $y_j=\Phi_j x_j\in \mathbb{R}^{d'}$ denote the low-dimensional measurement vector for pixel $j$ where $d'\leq d$ is size of the photosensor array in the compressive whisk-broom camera \cite{fow14}. As explained in \cite{single_pixel}, a DMD architecture can be used to produce a $\Phi_j$ with random entries in the range $[0,1]$ or random $\pm 1$ entries, resulting in a sub-Gaussian measurement matrix that satisfies the isometry conditions with a high probability \cite{simple}. Recall that the measurement matrix $\Phi_j$ is fixed in a FCA-based architecture while it can be distinct for each pixel in a DMD-based architecture. 

As noted in \cite{sp_cs}, the orthogonal projection onto the row-space of $\Phi_j$ can be computed as $P_j=\Phi_j^T (\Phi_j\Phi_j^T)^{-1}\Phi_j$. Consequently, an (unbiased) estimator for the inner product $x_j^T w$ (assuming a fixed $x_j$ and $w$) based on the compressive measurements would be $y_j^T (\Phi_j\Phi_j^T)^{-1} \Phi_j w$. As a result, the soft-margin SVM based on the compressive measurements can be expressed as:
\begin{equation}
\hat{w}^*=\arg\min_{w}  
\frac{1}{n}\sum_{j=1}^n\ell(z_j y_j^T (\Phi_j\Phi_j^T)^{-1} \Phi_j w)
+ \frac{\lambda}{2} \|w\|^2_2
\label{eq:svm_primal_rand}
\end{equation} 
(we have omitted the bias term $b$ for simplicity).

We must note that the formulation in (\ref{eq:svm_primal_rand}) is different from what was suggested in \cite{svm_cs} for a fixed measurement matrix. In particular, we solve for $\hat{w}^*$ in the $d$-dimensional space. Meanwhile, the methodology in \cite{svm_cs} would result in the following optimization problem:
\begin{equation}
\tilde{w}^*=\arg\min_{w} \frac{1}{n}\sum_{j=1}^n\ell(z_j y_j^T w) + \frac{\lambda}{2} \|w\|^2_2
\label{eq:svm_primal_rand2}
\end{equation} 
which solves for $\tilde{w}^*$ in the low-dimensional column-space of $\Phi$. Also note that, in the case of fixed measurement matrices, (\ref{eq:svm_primal_rand}) and (\ref{eq:svm_primal_rand2}) correspond to the same problem with the relationship $\hat{w}^*= \Phi^T (\Phi\Phi^T)^{-1} \tilde{w}^*$ (because of the $\ell_2$ regularization term which zeros the components of $\hat{w}^*$ which lie in the null-space of $\Phi$). In other words, (\ref{eq:svm_primal_rand}) represents a generalization of (\ref{eq:svm_primal_rand2}) for the case when the measurement matrices are not necessarily the same. This allows us to compare the two cases of $a$) having a fixed measurement matrix and $b$) having a distinct measurement matrix for each pixel, which is the subject of this paper. For simplicity, assume that each $\Phi_j$ consists of a subset of $d'$ rows from a random orthonormal matrix, or equivalently $\Phi_j\Phi_j^T=I_{d'}$; thus, $P_j=\Phi_j^T\Phi_j$. Also assume that, in the case of DMD-based sensing, each $\Phi_j$ is generated independently of the other measurement matrices.

Following the recent line of work in the area of randomized optimization, for example \cite{r19}, we refer to the new loss $\ell(z_j x_j^T \Phi_j^T (\Phi_j\Phi_j^T)^{-1} \Phi_j w)$ as the \textit{sketch} of the loss, or simply the \textit{sketched loss} to distinguish it from the true loss $\ell(z_j x_j^T w)$. Similarly, we refer to $\hat{w}^*$ as the \textit{sketched classifier} as opposed to the ground-truth classifier $w^*$. 


\begin{figure}[t]
\begin{minipage}[b]{.49\linewidth}
  \centering
  \centerline{\includegraphics[width=\linewidth,trim=0 10 10 290,clip]{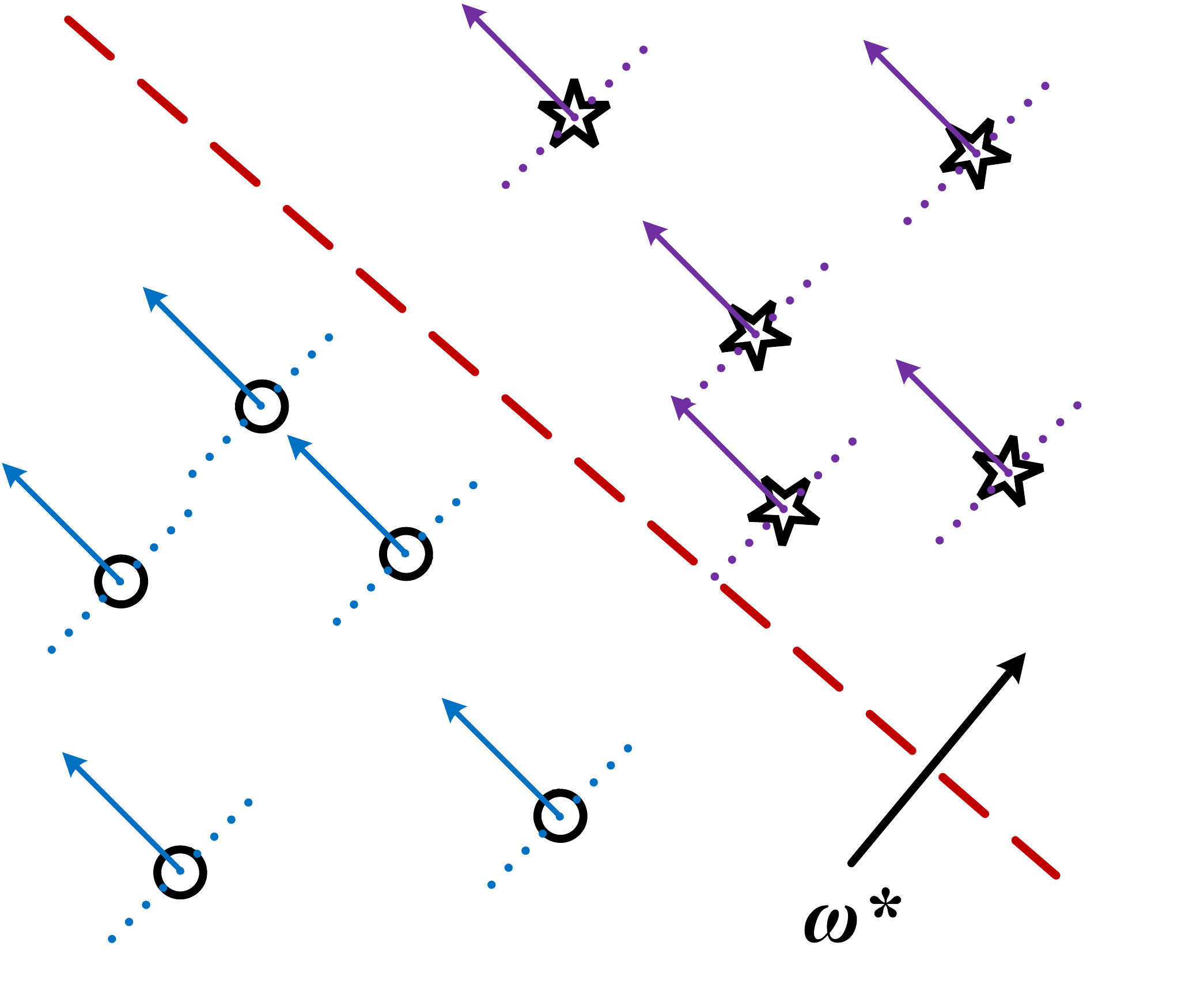}}
  \centerline{FCA-sensed data}
\end{minipage}
\hfill
\begin{minipage}[b]{0.49\linewidth}
  \centering
  \centerline{\includegraphics[width=\linewidth,trim=0 10 10 290,clip]{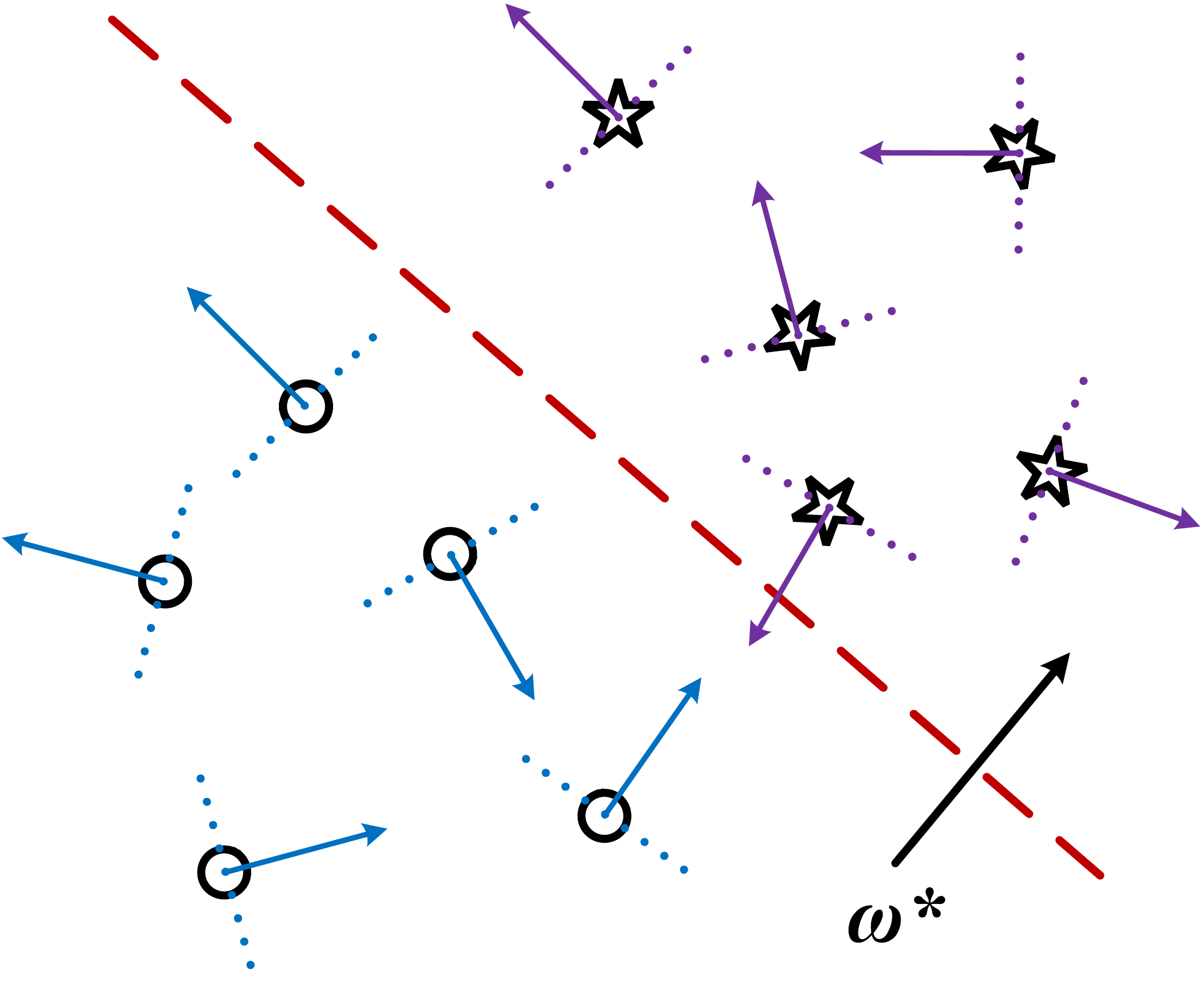}}
  \centerline{DMD-sensed data}
\end{minipage}
\caption{Linear SVM classification ---depicted for $d=2$ for illustration. Small arrows represent each 
$\Phi_j\in\mathbb{R}^{1\times d}$.}
\label{fig:F}
\end{figure}

Figure \ref{fig:F} depicts the two cases of using a fixed measurement matrix (FCA-sensed data) and distinct measurement matrices (DMD-sensed data) for training a linear classifier. It is helpful to imagine that, in the sketched problem, each $x_j$ is multiplied with $P_j w$ (the projection of $w$ onto the column-space of $\Phi_j$) since $y_j^T\Phi_j w=(P_j x_j)^T w= x_j^T (P_j w)$. 
As shown in Figure \ref{fig:F} (left) with $P_j=P$ for all $j$, there is a possibility that $w^*$ would nearly align with the null-space of the random low-rank matrix $P=\Phi^T\Phi$. For such $P$, any vector $P w$ may not well discriminate between the two classes and ultimately result in the classification failure. Figure \ref{fig:F} (right) depicts the case when a distinct measurement is used for each point. When $\Phi_j$ is symmetrically distributed in the space and $n$ is large, there is always a bunch of $\Phi_j$'s that nearly align with $w^*$ whereas other $\Phi_j$'s can be nearly orthogonal to $w^*$ or somewhere between the two extremes. This intuitive example hints about how measurement diversity pays off by making the optimization process more stable with respect to the variations in the random measurements and the separating hyperplane. 



\section{Simulations}
\label{sec:simo}

\subsection{Handling the bias term}

It is not difficult to see that employing a distinct $\Phi_j$ for each data vector $x_j$ necessitates having distinct values of bias $b_j$ (for each $\Phi_j$). Note that in the case of fixed measurement matrix, i.e. when $\Phi_j=\Phi$ for all $j$, bias terms would be all the same and linear SVM works normally as noted in \cite{svm_cs}. However, using a customized bias term for each point would clearly result in overfitting and the learned $\hat{w}^*$ would be of no practical value. Furthermore, the classifier cannot be used for prediction since the bias is unknown for the new input samples. In the following, we address these issues.

First, let $\mathcal{S}$ denote a set of $k$ distinct measurement matrices, i.e. $\mathcal{S}=\{\Phi^{(1)},\Phi^{(2)},\dots,\Phi^{(k)}\}$. Instead of using an arbitrary measurement matrix for each pixel, we draw an entry from $\mathcal{S}$ for each pixel. Given that $n\gg k$, each element of $\mathcal{S}$ is expected be utilized for more than once. This allows us to learn the bias for each outcome of measurement matrix (without the overfitting issue). Note that $k$ signifies the degree of measurement diversity: $k=1$ refers to the least diversity, i.e using a fixed measurement matrix, and measurement diversity is increased with $k$. The new optimization problem becomes:
\begin{eqnarray}
\left( \hat{w}^*, b_1^*,\dots,b_k^* \right) 
&=& \arg\min_{w,b_1,\dots,b_k} \frac{\lambda}{2} \|w\|^2_2 +  \nonumber\\
&& \frac{1}{n}\sum_{j=1}^n\ell(z_j y_j^T \Phi^{(t_j)} w + b_{t_j}) 
\label{eq:svm_khaf}
\end{eqnarray} 
where $t_j$ randomly (uniformly) maps each $j\in\{1,2,\dots,n\}$ to an element of $\{1,2,\dots,k\}$. The overfitting issue can now be restrained by tuning $k$; reducing $k$ results in less overfitting. In our simulations, we use $k\geq\ceil*{d/d'}$ to ensure that $\mathcal{S}$ spans $\mathbb{R}^d$ with a probability close to one.

For prediction, the corresponding bias term is selected from the set $\{b_1^*,b_2^*,\dots,b_k^*\}$. 

\subsection{Results}

The dataset used in this section is the well-known Pavia University dataset \cite{r18.1} which is available with the ground-truth labels\footnote{\url{http://www.ehu.eus/ccwintco/}}\footnote{The Indian Pines dataset was not included due to the small size of the image which is not sufficient for a large-scale cross-validation study.}. For each experiment, we perform a 2-fold cross-validation with $1000$ training and $1000$ testing samples. As discussed earlier, multi-categorical SVM classification algorithms typically rely on pair-wise or One-Against-One (OAO) classification results. Hence, we evaluate the sketched classifier on a OAO basis by reporting the pair-wise performances in a table . Finally, since the measurement operator is random and subject to variation in each experiment, we repeat each experiment for $1000$ times and perform a worst-case analysis of the results. 

Consider the case where a single measurement is made from each pixel, i.e. $d'=1$ and $\Phi_j\in\mathbb{R}^{1\times d}$ is a random vector in the $d$-dimensional spectral space. Clearly, this case represents an extreme scenario where the signal recovery would not be reliable and classification in the compressed domain becomes crucial, even at the receiver's side where the computational cost is not of greatest concern. For performance evaluation, we are interested in two aspects: ($a$) the prediction accuracy over the test dataset, ($b$) the recovery accuracy of the classifier (with respect to the ground-truth classifier) ---whose importance has been discussed in \cite{r21}.

\begin{figure}
\begin{minipage}[b]{.49\linewidth}
  \centering
  \centerline{\includegraphics[width=\linewidth,trim=0 10 10 390,clip]{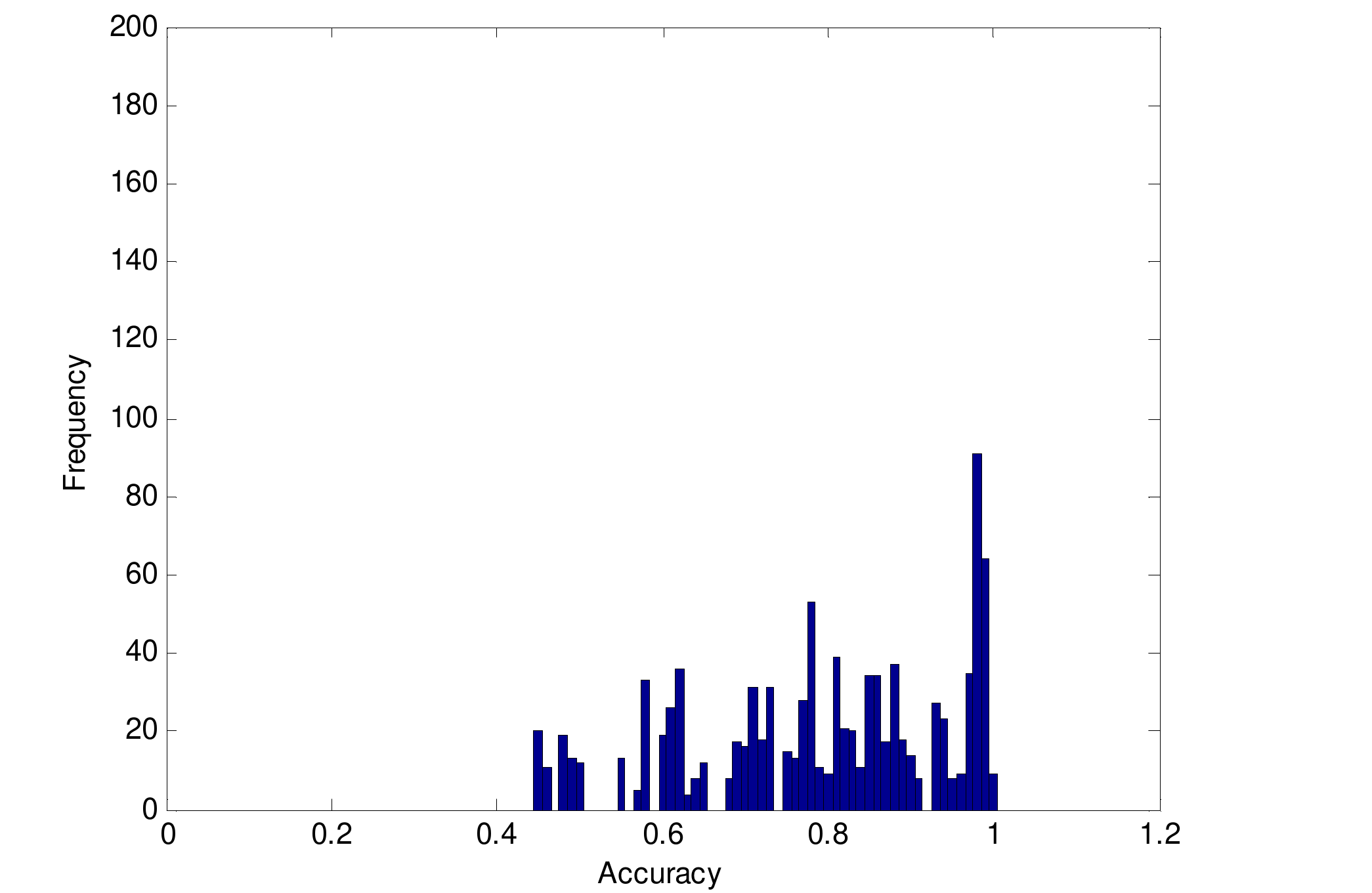}}
  \centerline{FCA measurement}
\end{minipage}
\hfill
\begin{minipage}[b]{0.49\linewidth}
  \centering
  \centerline{\includegraphics[width=\linewidth,trim=0 10 10 390,clip]{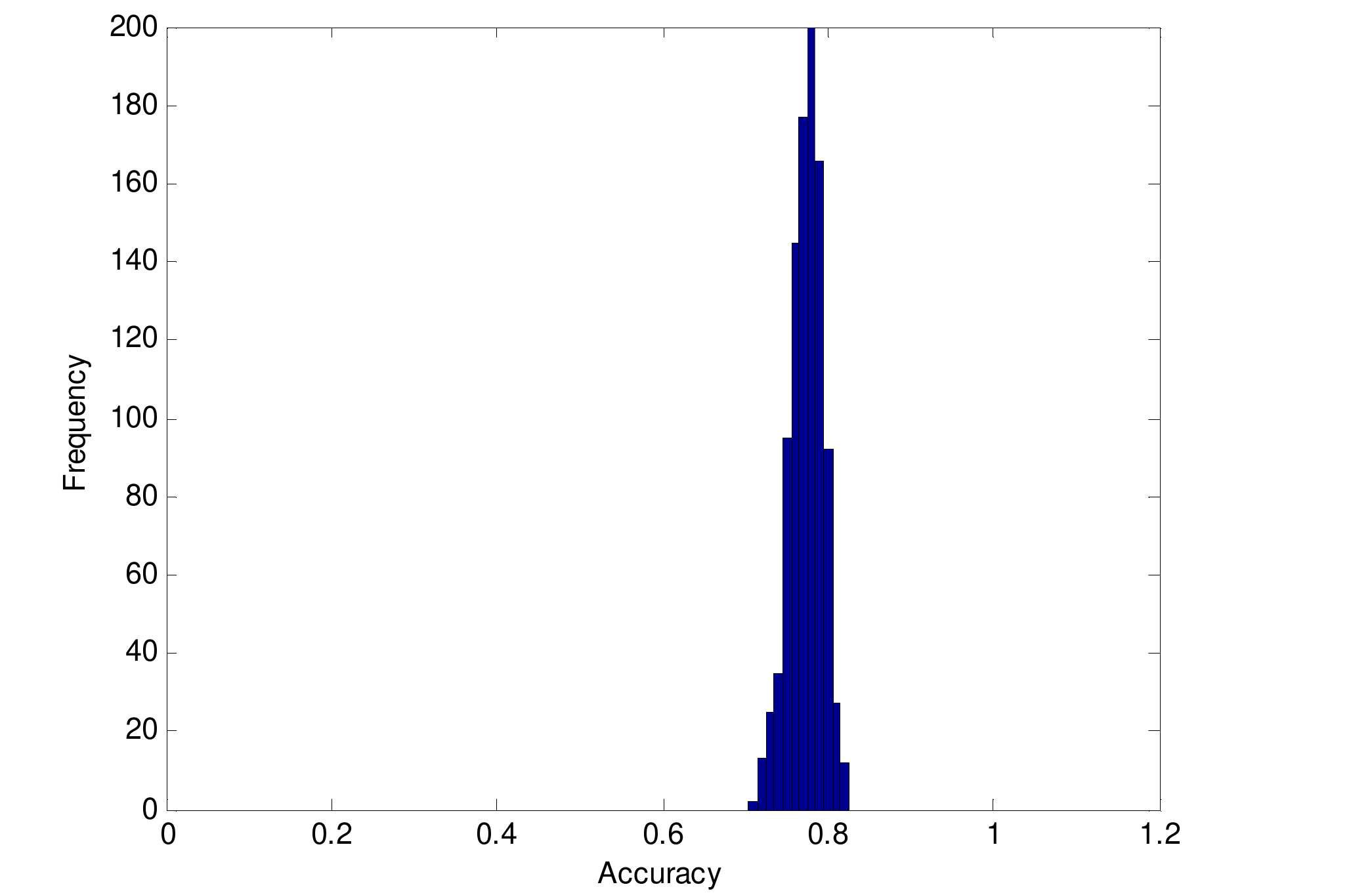}}
  \centerline{DMD measurement}
\end{minipage}
\caption{Distributions of the classification accuracy (Asphalt vs. Meadows) for the Pavia University dataset ($d'=1$).}
\label{fig:A}
\end{figure}

We define the classification accuracy as the minimum (worst) of the True Positive Rate (sensitivity) and the True Negative Rate (specificity). Figure \ref{fig:A} shows an instance of the distribution of the classification accuracy for a pair of classes over $1000$ random trials. As it can be seen, in the presence of measurement diversity, classification results are more consistent (reflected in the low variance of accuracy). Due to the limited space, we only report the worst-case OAO accuracies (i.e. the minimum pair-wise accuracies among $1000$ trials) for the Pavia scene. The results for the case of one-measurement-per-pixel ($d'=1$) are shown in Tables \ref{tab:pavia_fca} and \ref{tab:pavia_dmd}. Similarly, the results for the case of $d'=3$ (which is equivalent to the sampling rate of a typical RGB color camera) are shown in Tables \ref{tab:pavia_fca_3} and \ref{tab:pavia_dmd_3}. Note that the employed SVM classifier is linear and would not result in perfect accuracy (i.e. accuracy of one) when the classes are not linearly separable. To see this, we have reported ground-truth accuracies in Table \ref{tab:pavia_gt}.

\begin{table}
\caption{One FCA measurement per pixel: worst-case classification accuracies (1000 trials) for the Pavia scene.}
\label{tab:pavia_fca}
\centering
\begin{tabular}{|l|c|c|c|c|c|}\hline
Classes & Meadow & Gravel & Trees 	& Soil 	& Bricks \\\hline
Asphalt & 0.45 	& 0.38 	& 0.42 	& 0.36 	& 0.44 	\\\hline
Meadow 	&  			& 0.48 	& 0.48 	& 0.41 	& 0.47 	\\\hline
Gravel 	&  			&  			& 0.44	& 0.44 	& 0.44 	\\\hline
Trees 	&  			&  			&  			& 0.42 	& 0.53 	\\\hline
Soil 		&  			&  			&  			&  			& 0.44 	\\\hline
\end{tabular}
\end{table}

\begin{table}
\caption{One DMD measurement per pixel: worst-case classification accuracies (1000 trials) for the Pavia scene.}
\label{tab:pavia_dmd}
\centering
\begin{tabular}{|l|c|c|c|c|c|}\hline
Classes & Meadow & Gravel & Trees 	& Soil 	& Bricks \\\hline
Asphalt & 0.71 	& 0.64 	& 0.79 	& 0.60 	& 0.71 	\\\hline
Meadow 	&  			& 0.72 	& 0.61 	& 0.46 	& 0.73 	\\\hline
Gravel 	&  			&  			& 0.79	& 0.60 	& 0.44 	\\\hline
Trees 	&  			&  			&  			& 0.69 	& 0.79 	\\\hline
Soil 		&  			&  			&  			&  			& 0.60 	\\\hline
\end{tabular}
\end{table}

\begin{table}
\caption{Three FCA measurements per pixel: worst-case classification accuracies (1000 trials) for the Pavia scene.}
\label{tab:pavia_fca_3}
\centering
\begin{tabular}{|l|c|c|c|c|c|}\hline
Classes & Meadow & Gravel & Trees 	& Soil 	& Bricks \\\hline
Asphalt & 0.61 	& 0.80 	& 0.94 	& 0.63 	& 0.86 	\\\hline
Meadow 	&  			& 0.67 	& 0.82 	& 0.50 	& 0.62 	\\\hline
Gravel 	&  			&  			& 0.94	& 0.62 	& 0.54 	\\\hline
Trees 	&  			&  			&  			& 0.89 	& 0.93 	\\\hline
Soil 		&  			&  			&  			&  			& 0.66 	\\\hline
\end{tabular}
\end{table}

\begin{table}
\caption{Three DMD measurements per pixel: worst-case classification accuracies (1000 trials) for the Pavia scene.}
\label{tab:pavia_dmd_3}
\centering
\begin{tabular}{|l|c|c|c|c|c|}\hline
Classes & Meadow & Gravel & Trees 	& Soil 	& Bricks \\\hline
Asphalt & 0.91 	& 0.76 	& 0.96 	& 0.87 	& 0.84 	\\\hline
Meadow 	&  			& 0.90 	& 0.82 	& 0.57 	& 0.91 	\\\hline
Gravel 	&  			&  			& 0.95	& 0.82 	& 0.49 	\\\hline
Trees 	&  			&  			&  			& 0.93 	& 0.96 	\\\hline
Soil 		&  			&  			&  			&  			& 0.80 	\\\hline
\end{tabular}
\end{table}

\begin{table}
\caption{Ground-truth accuracies for the Pavia scene.}
\label{tab:pavia_gt}
\centering
\begin{tabular}{|l|c|c|c|c|c|}\hline
Classes & Meadow & Gravel & Trees 	& Soil 	& Bricks \\\hline
Asphalt & 1.00 	& 0.97 	& 0.97 	& 1.00 	& 0.94 	\\\hline
Meadow 	&  			& 0.99 	& 0.96 	& 0.89 	& 0.99 	\\\hline
Gravel 	&  			&  			& 1.00	& 1.00 	& 0.86 	\\\hline
Trees 	&  			&  			&  			& 0.98 	& 1.00 	\\\hline
Soil 		&  			&  			&  			&  			& 0.99 	\\\hline
\end{tabular}
\end{table}

To measure the classifier recovery accuracy, we compute the cosine similarity, or equivalently the correlation, between $\hat{w}^*$ and $w^*$:
$$C(\hat{w}^*,w^*)=\frac{\left<\hat{w}^*,w^*\right>}{\|\hat{w}^*\|_2 \|w^*\|_2} $$
In Tables \ref{tab:pavia_err_fca_3} and \ref{tab:pavia_err_dmd_3}, we have reported the average recovery accuracy for the case of three-measurements-per-pixel (i.e. $d'=3$). 

%

\begin{table}
\caption{Three FCA measurements per pixel: average recovery accuracy (1000 trials) for the Pavia scene.}
\label{tab:pavia_err_fca_3}
\centering
\begin{tabular}{|l|c|c|c|c|c|}\hline
Classes & Meadow & Gravel & Trees 	& Soil 	& Bricks \\\hline
Asphalt & 0.051 & 0.055 & 0.113 & 0.056 & 0.048 	\\\hline
Meadow 	&  			& 0.100 & 0.033 & 0.019 & 0.077 	\\\hline
Gravel 	&  			&  			& 0.122	& 0.064 & 0.050 	\\\hline
Trees 	&  			&  			&  			& 0.017 & 0.123 	\\\hline
Soil 		&  			&  			&  			&  			& 0.031 	\\\hline
\end{tabular}
\end{table}

\begin{table}
\caption{Three DMD measurements per pixel: average recovery accuracy (1000 trials) for the Pavia scene.}
\label{tab:pavia_err_dmd_3}
\centering
\begin{tabular}{|l|c|c|c|c|c|}\hline
Classes & Meadow & Gravel & Trees 	& Soil 	& Bricks \\\hline
Asphalt & 0.164 & 0.189 & 0.483 & 0.129 & 0.132 	\\\hline
Meadow 	&  			& 0.468 & 0.147 & 0.140 & 0.380 	\\\hline
Gravel 	&  			&  			& 0.617	& 0.272 & 0.197 	\\\hline
Trees 	&  			&  			&  			& 0.102 & 0.582 	\\\hline
Soil 		&  			&  			&  			&  			& 0.128 	\\\hline
\end{tabular}
\end{table}

\vspace{-10pt}
\section{Conclusion}
\label{sec:conclusion}

In the field of ensemble learning, it has been discovered that the diversity among the base learners enhances the overall learning performance \cite{r20}. Meanwhile, our aim has been to exploit the diversity that can be efficiently built into the sensing system. Both measurement schemes of pixel-invariant (measurement without diversity) and pixel-varying (measurement with diversity) have been suggested as practical designs for compressive hyperspectral cameras \cite{fow14}. The presented analysis indicates that employing a DMD would result in more accurate recovery of the classifier and a more stable classification performance compared to the case when an FCA is used. Meanwhile, for tasks that only concern class prediction (and not the recovery of the classifier), FCA is (on average) a suitable low-cost alternative to the DMD architecture.

\bibliographystyle{IEEEbib}

\AtEndEnvironment{thebibliography}{

\bibitem{r10}
W.K. Ma, J.M. Bioucas Dias, Tsung Han Chan, N. Gillis, P. Gader, A.J. Plaza, A. Ambikapathi and Chong-Yung Chi, ``A Signal Processing Perspective on Hyperspectral Unmixing: Insights from Remote Sensing,'' \emph{Signal Processing Magazine, IEEE}, vol.31, no.1, pp.67,81, January 2014.

\bibitem{r11}
F. Melgani and L. Bruzzone, ``Classification of hyperspectral remote sensing images with support vector machines,'' \emph{IEEE Transactions on Geoscience and Remote Sensing}, vol. 42, no. 8, pp. 1778–1790, August 2004.

\bibitem{r12}
O. Chapelle, ``Training a support vector machine in the primal,'' \emph{Neural Computing}, vol. 19(5), pp. 1155–1178, 2007.

\bibitem{r17}
This dataset was gathered by AVIRIS sensor over the Indian Pines test site in North-western Indiana and consists of $145\times145$ pixels and 224 spectral reflectance bands in the wavelength range 0.4 to 2.5e-6 meters. 


\bibitem{r18.1}
This scene was acquired by the ROSIS sensor during a flight campaign over Pavia, northern Italy. The number of spectral bands is 103 and the spatial resolution is $610\times 610$ pixels. Ground-truth consists of 9 classes.

\bibitem{r19}
M. Pilanci, Martin J. Wainwright, ``Randomized Sketches of Convex Programs with Sharp Guarantees,'' arXiv: 1404.7203 [cs.IT], April 2014.

\bibitem{r20}
B. Waske, S. Van Der Linden, J.A. Benediktsson, A. Rabe and P. Hostert, ``Sensitivity of support vector machines to random feature selection in classification of hyperspectral data,'' \emph{IEEE Transactions on Geoscience and Remote Sensing}, vol. 48, pp. 2880–2889, 2010.


}

\bibliography{refs}
\end{document}